\pdfoutput=1

\documentclass[11pt]{article}

\usepackage[]{EMNLP2023}

\usepackage{times}
\usepackage{latexsym}

\usepackage[T1]{fontenc}

\usepackage[utf8]{inputenc}

\usepackage{microtype}

\usepackage{inconsolata}

\usepackage{graphicx}
\usepackage{float}
\usepackage{subcaption}
\usepackage{multirow}
\usepackage{makecell}
\usepackage{amsmath}
\usepackage{multicol}
\usepackage{booktabs}
\usepackage{enumitem}

\usepackage{xcolor}         
\usepackage{colortbl}         
\usepackage{amssymb}  

\newcommand{\factor}[1]{{\fontfamily{cmtt}\selectfont #1}}
\title{The ICL Consistency Test
}

\author{Lucas Weber \\
  University Pompeu Fabra \\
  \texttt{lucas.weber@upf.edu} \\\And
  Elia Bruni \\
  Osnabrück University \\
  \texttt{elia.bruni@gmail.com} \\\And
  Dieuwke Hupkes \\
  FAIR \\
  \texttt{dieuwkehupkes@meta.com}
}

\begin{document}
\maketitle
\begin{abstract}
Just like the previous generation of \textit{task-tuned} models, large language models (LLMs) that are adapted to tasks via prompt-based methods like \emph{in-context-learning} (ICL) perform well in some setups but not in others. 
This lack of consistency in prompt-based learning hints at a lack of robust generalisation.
We here introduce the \textit{ICL consistency test} -- a contribution to the GenBench collaborative benchmark task (CBT) -- which evaluates how consistent a model makes predictions across many different setups while using the same data.
The test is based on different established natural language inference tasks.
We provide preprocessed data constituting 96 different `setups' and a metric that estimates model consistency across these setups. 
The metric is provided on a fine-grained level to understand what properties of a setup render predictions unstable and on an aggregated level to compare overall model consistency.
We conduct an empirical analysis of eight state-of-the-art models, and our consistency metric reveals how all tested LLMs lack robust generalisation.

\end{abstract}

\section{Introduction}
\label{sec:intro}
Prompting-based approaches that use in-context learning \citep[from here on \emph{ICL};][]{brown2020language} such as few-shot \citep{radford2019language} or zero-shot \citep{wei2021finetuned} inference have recently superseded task-specific parameter tuning as the go-to method to adapt pre-trained language models to any task of interest. 

Prompt-based task adaptation has the benefit that it eliminates the need for costly, task-specific fine-tuning and provides greater flexibility, as a single model can be applied to many tasks without further tuning.
Previous research also suggests that out-of-distribution generalisation is less of a problem for prompt-based learners \citep{awadalla2022exploring, si2023prompting}.
However, ICL currently yields overall weaker performance compared to task-tuning and is less stable and reliable on many common benchmarks \citep[see e.g.][]{bang2023multitask, ohmer2023evaluating, min2022rethinking,lu2021fantastically,zhao2021calibrate}.
If we change the setup in which a model is prompted -- for example, by changing the instructions that explain the task -- model predictions can change unpredictably, even if the change is irrelevant to the task at hand \citep{liang2022holistic}.
This chaotic model behaviour is a problem in real-world applications and reveals a more profound problem of non-robust generalisation.
A model that generalises according to the underlying task distribution should be invariant to changes in the setup that do \emph{not} change the nature of the task. 

The ICL consistency test presented in this paper evaluates the ability of a model to make consistent predictions for a data point when presented across many different setups. 
In other words, it tests the ability to generalise \emph{robustly} and \emph{across tasks} under an assumed shift between the pretraining data and the data that we test on \citep[compare GenBench; ][]{hupkes2023taxonomy}.
For most large language models (LLMs), we have neither insight nor control over the training data.
However, we can assume that only a neglectable amount of the pretraining data follow a similar format to our prompts and require classification.
We further assume that ICL latches on irrelevant properties in the input data. 
We classify the shifts in our data as \textit{assumed shifts}.
The test is a contribution to the Collaborative Benchmarking Task \citep[CBT;][]{hupkes2023taxonomy} and can be located in the GenBench taxonomy as shown in Table~\ref{tab:GenBenchEvalCard}.

\newcommand{\tabularwidth}{\textwidth}

\newcommand{\expone}{$\square$}      

\begin{table*}[h!]
\centering
\renewcommand{\arraystretch}{1.1}         
\setlength{\tabcolsep}{0mm}         
\begin{tabular}{|p{\tabularwidth}<{\centering}|}         
\hline
               
\rowcolor{gray!60}               
\textbf{Motivation} \\               
\footnotesize
\begin{tabular}{p{0.25\tabularwidth}<{\centering} p{0.25\tabularwidth}<{\centering} p{0.25\tabularwidth}<{\centering} p{0.25\tabularwidth}<{\centering}}                        
\textit{Practical} & \textit{Cognitive} & \textit{Intrinsic} & \textit{Fairness}\\
\expone\hspace{0.8mm}		
& 		
& \expone\hspace{0.8mm}		
& 		

\vspace{2mm} \\
\end{tabular}\\
               
\rowcolor{gray!60}               
\textbf{Generalisation type} \\               
\footnotesize
\begin{tabular}{m{0.21\tabularwidth}<{\centering} m{0.2\tabularwidth}<{\centering} m{0.13\tabularwidth}<{\centering} m{0.13\tabularwidth}<{\centering} m{0.13\tabularwidth}<{\centering} m{0.19\tabularwidth}<{\centering} m{0.01\tabularwidth}<{\centering}}                   
\textit{Compositional} & \textit{Structural} & \textit{Cross Task} & \textit{Cross Language} & \textit{Cross Domain} & \textit{Robustness} & \\
& 		
& \expone\hspace{0.8mm}		
& 		
& 		
& \expone\hspace{0.8mm}		
&
\end{tabular}\\
             
\rowcolor{gray!60}             
\textbf{Shift type} \\             
\footnotesize
\begin{tabular}{p{0.25\tabularwidth}<{\centering} p{0.25\tabularwidth}<{\centering} p{0.25\tabularwidth}<{\centering} p{0.24\tabularwidth}<{\centering} p{0.01\tabularwidth}<{\centering}}                        
\textit{Covariate} & \textit{Label} & \textit{Full} & \textit{Assumed} & \\  
& 		
& 		
& \expone\hspace{0.8mm}		
&
\vspace{2mm} \\
\end{tabular}\\
             
\rowcolor{gray!60}             
\textbf{Shift source} \\             
\footnotesize
\begin{tabular}{p{0.25\tabularwidth}<{\centering} p{0.25\tabularwidth}<{\centering} p{0.25\tabularwidth}<{\centering} p{0.25\tabularwidth}<{\centering}}                          
\textit{Naturally occuring} & \textit{Partitioned natural} & \textit{Generated shift} & \textit{Fully generated}\\
& \expone\hspace{0.8mm}		
& 		
& 		

\vspace{2mm} \\
\end{tabular}\\
             
\rowcolor{gray!60}             
\textbf{Shift locus}\\             
\footnotesize
\begin{tabular}{p{0.25\tabularwidth}<{\centering} p{0.25\tabularwidth}<{\centering} p{0.25\tabularwidth}<{\centering} p{0.24\tabularwidth}<{\centering}p{0.01\tabularwidth}<{\centering}}                         
\textit{Train--test} & \textit{Finetune train--test} & \textit{Pretrain--train} & \textit{Pretrain--test} &\\
& 		
& 		
& \expone\hspace{0.8mm}		
&
\end{tabular}\\

\hline
\end{tabular}
\caption{The GenBench Evaluation card corresponding to the ICL consistency test.}
\label{tab:GenBenchEvalCard}
\end{table*}

The paper is structured in the following way:
First, we provide background information on in-context learning and inconsistency of predictions in prompt-based model adaptation (Section~\ref{sec:background}). 
After that, we introduce our task by laying out its motivation (\ref{subsec:motivation}), document the used data (\ref{subsec:data}), and then describe how we construct the test (\ref{subsec:setups_and_factors}) as well as the used metrics (\ref{subsec:metrics}).
Ultimately, we empirically test the ICL consistency test on eight different models, showing how state-of-the-art LLMs perform surprisingly poorly on our benchmark, promising that the ICL consistency test can be an important and complementing marker of generalisation besides regular accuracy scores.

\section{Background}
\label{sec:background}

In-context learning (ICL) describes the adaptation of a model by inferring a task from the left-handed context of a tested input to generate a matching output.
ICL is divisible into two categories: (1) few-shot learning, which involves providing a limited set of examples (comprising input-output pairs) in the left context of a tested input, and (2) zero-shot learning, which pertains to situations without any examples.
The ICL consistency test uses the few-shot settings.

In contrast to task-tuning, ICL is a considerably cheaper adaptation method as it does not require any parameter updates. 
\citet{akyurek2022learning} and \citet{garg2022can} show that transformer adaptation via ICL exhibits sufficient expressivity to realise simple linear algorithms, small neural networks or decision trees. 
Although ICL naturally arises as untuned LLMs grow in size, as noted by \cite{brown2020language}, these \emph{`vanilla}' LLMs often fall short in performance compared to the fine-tuned state-of-the-art models on common NLP benchmarks, as shown by \citep{liang2022holistic}.
Further, ICL is highly unstable: previous research has shown how the order of in-context examples \citep{lu2021fantastically}, the recency of certain labels in the context \citep{zhao2021calibrate} or the format of the prompt \citep{mishra2021reframing} as well as the distribution of training examples and the label space \citep{min2022rethinking} strongly influence the model's predictions.
Curiously, the correspondence of inputs with their labels is less important \citep{min2022rethinking}.
Further, \citet{kim2022ground} paint an even more differentiated picture, demonstrating that in-context input-label mapping \emph{does} matter, but that it depends on other factors such as model size or instruction verbosity. 
Work of \citet{wei2023larger} goes similarly, showing that in-context learners can acquire new semantically non-sensical mappings from in-context examples if presented in the correct setup.
Similar to the robustness of task-finetuned models \citep[for examples, see][]{hupkes2023taxonomy}, in-context learning appears to be influenced by certain factors in the setup that are not relevant to the task at hand. 
We can further deduce from previous research that the reasons for inconsistency across prompting setups are not straightforward, and benchmarks to monitor progress are needed.

\section{Task description}
\label{sec:task}

We here introduce the ICL consistency test that evaluates a model's ability to make consistent predictions on the same data point, independent of the respective evaluation setup. 
It compares a model's prediction across many different prompting \textit{setups}.

We define \textit{setups} through the presence or absence of different binary \emph{factors}, which are simple --- usually binary --- choices in the prompt design (e.g. do I use instruction A or B to prompt the model).
The ICL consistency test provides preprocessed prompts for all possible combinations of factors (i.e. for all possible setups) for the ANLI and MNLI datasets \citep{nie2019adversarial, williams2017broad}. Further, we use freely available instruction templates \citep[promptsource (P3);][]{bach2022promptsource} to preprocesses to compose our setups.

In the following, we will depict the task in more detail:
We start by describing the issue of inconsistent predictions in in-context learners (Section~\ref{subsec:motivation}), to then introduce our task with greater detail, presenting the used data (Section~\ref{subsec:data}), the characteristics of the employed setups (Section~\ref{subsec:setups_and_factors}), and ultimately, the evaluation metric introduced to estimate a model's prediction consistency (Section~\ref{subsec:metrics}).

\subsection{Motivation}
\label{subsec:motivation}
Consistency measures are complementary to accuracy: imagine a scenario in which a model is evaluated with two different but equally valid setups. 
For example, one could query a model for the sentiment of a sentence \factor{<x>} using either of the following instructions:

\begin{description}
    \item[Instruction 1] Please state whether the following sentence is positive, negative, or neutral: \factor{<x>}
    \item[Instruction 2] Given the sentence: "\factor{<x>}", please classify its sentiment as positive, negative, or neutral.
\end{description}

\noindent While both prompts are superficially different, their conveyed query is exactly the same.
Let's assume the model predicts the same proportion of correct labels in either setup but does so on a different subset of the evaluation data.
The accuracy score has the same value in either setting and, therefore, could let us assume that we must improve the model's ability to solve the task at hand.
In reality, however, the main issue is the model's questionable generalisation and lack of robustness to irrelevant changes in the prompt.
We have seen in the background section that prompt-based learners lack this type of robustness more often than not. 
To accurately analyse errors, it is therefore crucial to have a tool to estimate reliability by systematically evaluating a model's consistency.
Additionally, a consistency benchmark is crucial to estimating whether new methods improve model behaviour robustly across many conditions or are only successful in a particular setting.
For these reasons, we introduce the ICL consistency test.

\subsection{Data}
\label{subsec:data}
We use freely available and established data sources to construct the ICL consistency test.

\paragraph{Instructions}
Instructions explain in natural language to a model which task it should solve and wrap the original input $x$ from a given data set.
We source natural language instructions from different subsets of the crowdsourced \emph{promptsource templates} \citep[from here on `P3';][for examples, see Appendix~\ref{subapp:p3_examples}]{bach2022promptsource}, with the exact template being used depending on the specific setup that is evaluated.
Exact information on which instructions are employed is given in Section~\ref{subsec:setups_and_factors}.

\paragraph{Data}
We use the p3 instructions templates to wrap data points from the ANLI \citep{nie2019adversarial} and MNLI \citep{williams2017broad} datasets. 
For each of the datasets, we randomly draw a subset of 600 data points from the respective validation sets, and -- in the case of ANLI -- we draw to equal parts from the validation sets of its three distinct subsets.
We give each data point a unique data ID.
The ANLI and MNLI are implemented as different subtasks in the GenBench code submission.

\paragraph{In-context examples}
We provide solved examples in the left-handed context of the model input as an aid for the model to infer the task it has to solve \citep[as done in][]{brown2020language}. 
These in-context examples are constructed similarly to the target examples but have their ground truth label concatenated.
To select in-context examples, we randomly draw data points from the respective full training sets.
The label space, the instructions, the number or even the task of in-context examples can, again, differ depending on the evaluated setup.

Examples of prompts can be found in Appendix~\ref{subapp:p3_examples}.
\begin{figure*}[h!]
  \centering
    \begin{subfigure}[h]{0.345\linewidth}
    \includegraphics[width=\linewidth]{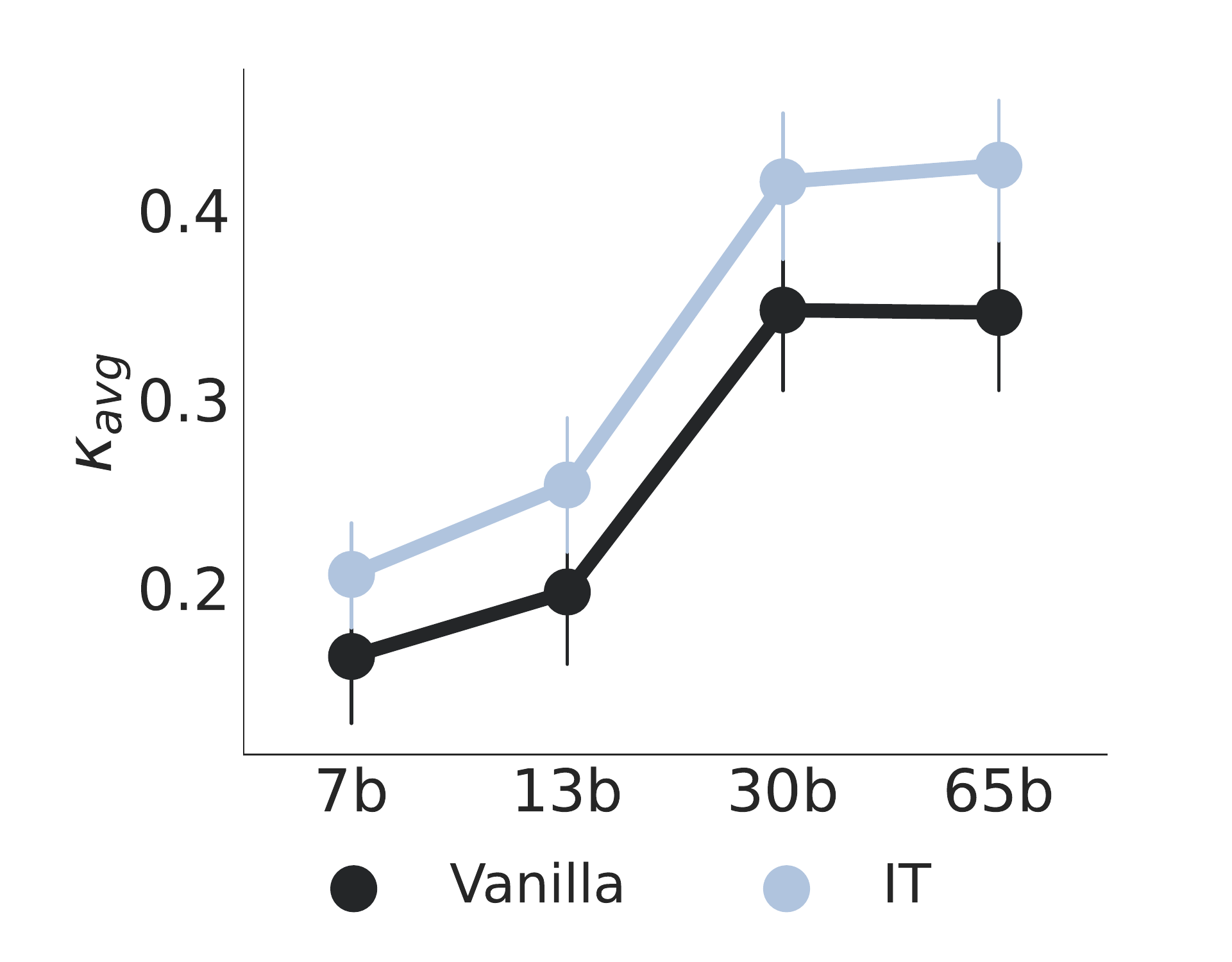}
    \caption{}
    \label{subfig:consistency_templates_per_model}
  \end{subfigure}
  \begin{subfigure}[h]{0.315\linewidth}
    \includegraphics[width=\linewidth]{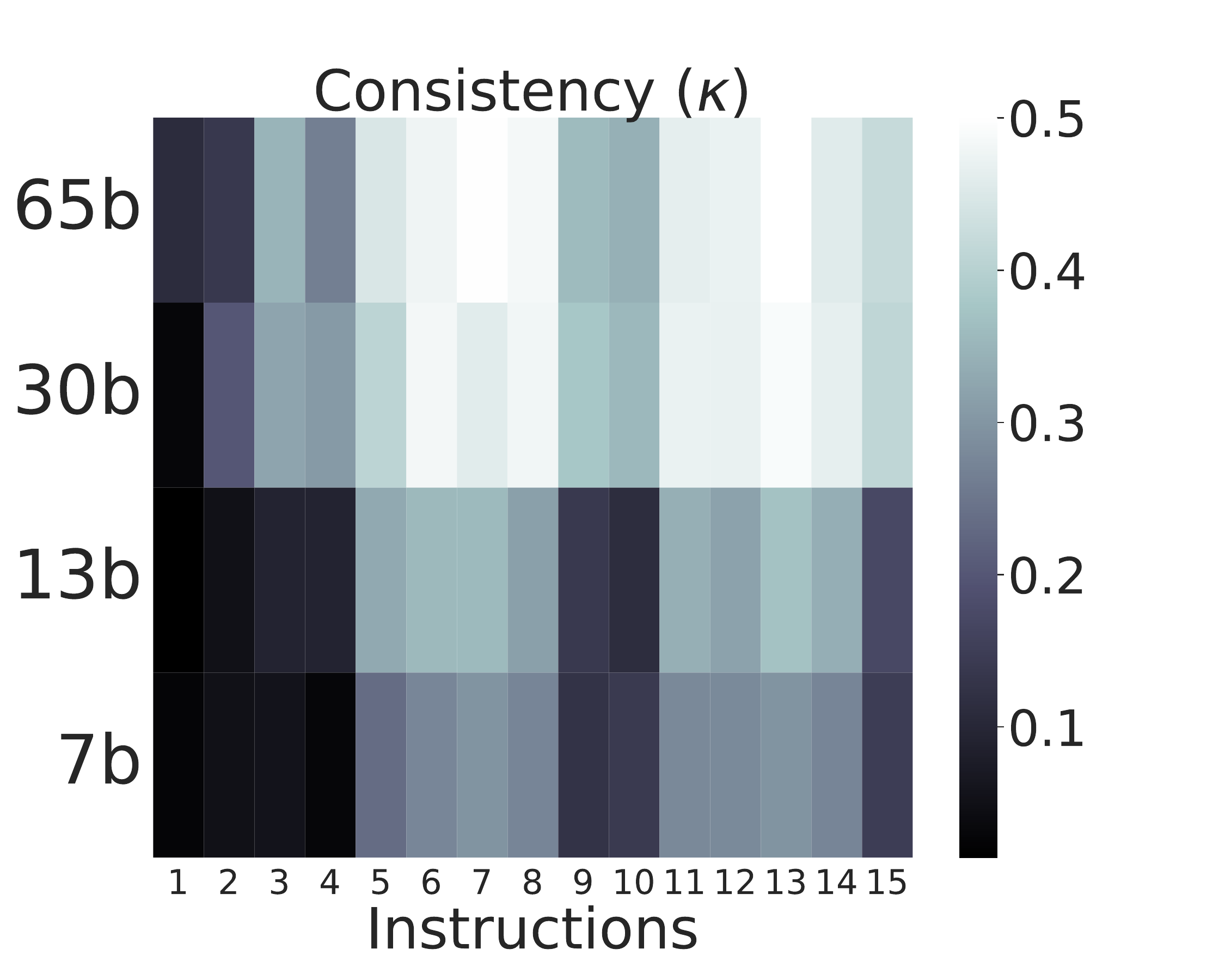}
    \caption{}
    \label{subfig:consistency_templates_per_template}
  \end{subfigure}
  \begin{subfigure}[h]{0.315\linewidth}
    \includegraphics[width=\linewidth]{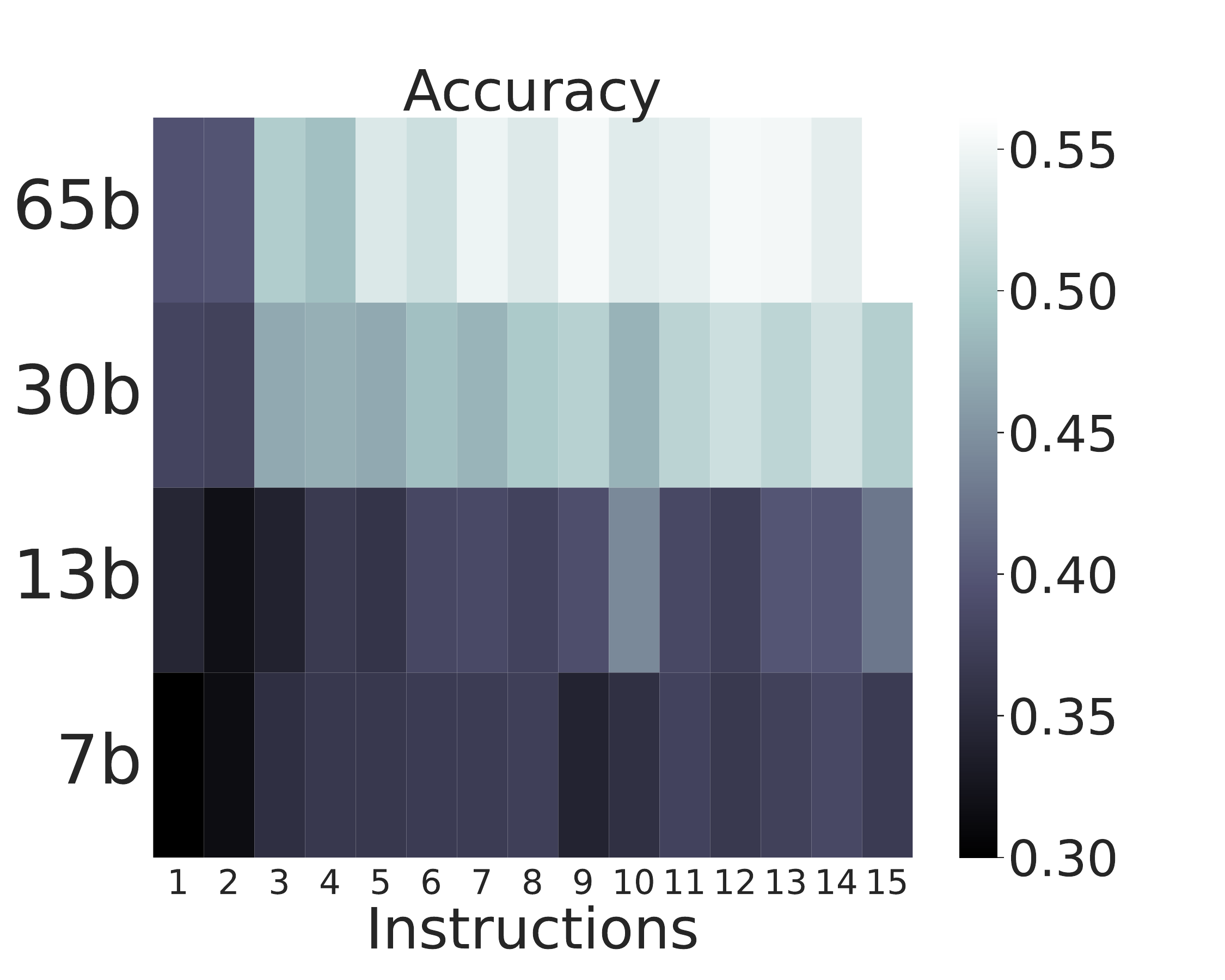}
    \caption{}
    \label{subfig:heatmap_performance_templates}
  \end{subfigure}
  \caption{Figure (a) shows the consistency of a model when used with all 15 different P3 instructions, in an otherwise fixed setup. A value of 1 indicates high consistency, and a value of 0 indicates low consistency; 
  Figure (b) shows how consistent individual instructions are with all other instructions. A value of 0 indicates a complete change of predictions while a value of 1 indicates perfect agreement; Figure (c) shows the respective accuracies of the instructions in Figure (b).}
  \label{fig:diversity_predictions}
\end{figure*}

\subsection{Setups and factors}
\label{subsec:setups_and_factors}
We estimate a model's robustness by evaluating its prediction's consistency on the same data point across many different setups. 
We define each setup through the absence or presence of each of a range of binary factors $\lambda$. 

\subsubsection{Description of factors} 
\label{subsubsec:description_factors}
We include the following $\lambda$s in our test\footnote{For more detailed explanations on the different factors and the respective motivation to include them, we refer to Appendix~\ref{app:factors_details}}:
    \begin{description}[noitemsep]
    \item[n-shots] We use many ($k$ = 5) or few ($k$ = 2) in-context examples in the prompt.
    \item[High-performing (HP) instructions] We use two groups of semantically equivalent but \emph{differently} performing instruction templates (high- vs. low-performing; more details in Section~\ref{subsubsec:probing_templates}).
    \item[Balanced labels] We use examples with labels that are balanced across all possible classes in the context, or we use randomly sampled examples.
    \item[Cross-templates] We draw in-context instructions randomly from all available instruction templates, or we use the same instructions as for the target.
    \item[Cross-task] We use samples from another classification task \citep[QQP;][]{wang2017bilateral} as in-context examples or samples from the same task as the target task (ANLI / MNLI).
    \item[Instructions] We use different semantically equivalent target instructions that perform \emph{similarly} (more details in Section~\ref{subsubsec:probing_templates}).
    \item[One label] We use only in-context examples with a single randomly selected label, or we use randomly selected in-context examples.
    \end{description}
    
\noindent Arranging the above factors in all possible combinations results in 96 setups.
Combining the 96 setups with our randomly sampled 600 data points results in \textbf{57\_600 samples} for each subtask.
Each setup has a unique setup ID, which can uniquely identify a specific prompt when combined with a respective data ID (details on the composition of setup IDs can be found in Appendix~\ref{app:setup_IDs}).
Besides the provided factors, it is also possible to augment the ICL consistency test with additional factors using the code implementation submitted to the GenBench CBT (for details, see Appendix~\ref{app:custom_factors}).

\subsection{Metrics}
\label{subsec:metrics}

\paragraph{Cohen's kappa}
We measure the consistency of model predictions using Cohen's $\kappa$ \citep[][]{cohen1960coefficient}, a measure of interrater agreement adjusted for agreement by chance.
The metric $\kappa$ equals 1 if two (or more) sets of predictions perfectly align while agreement by chance results in $\kappa$ equalling 0. 
In our case, we calculate $\kappa$ to compare the predictions of a model before and after we change the value of a factor $\lambda$ across all possible setups.
For example, we take the predictions from all setups in which in-context examples have the same label (the factors \factor{one label} is present) and compare it to the case in which we have different labels for the in-context examples (the factors \factor{one label} is absent). 
With all other factors being constant, we can estimate how much this factor changed the model prediction (or, inversely, how robust a model is) by calculating $\kappa$.

To ensure meaningful scores, we mask out all predictions that are not within the label distribution of the respective task. 
Finally, we get the overall model consistency $\kappa_{avg}$, by averaging across the $\kappa$ values of all factors.


\begin{figure*}[h!]
  \centering
  \begin{subfigure}[h]{0.99\linewidth}
  \centering
     \includegraphics[width=\linewidth]{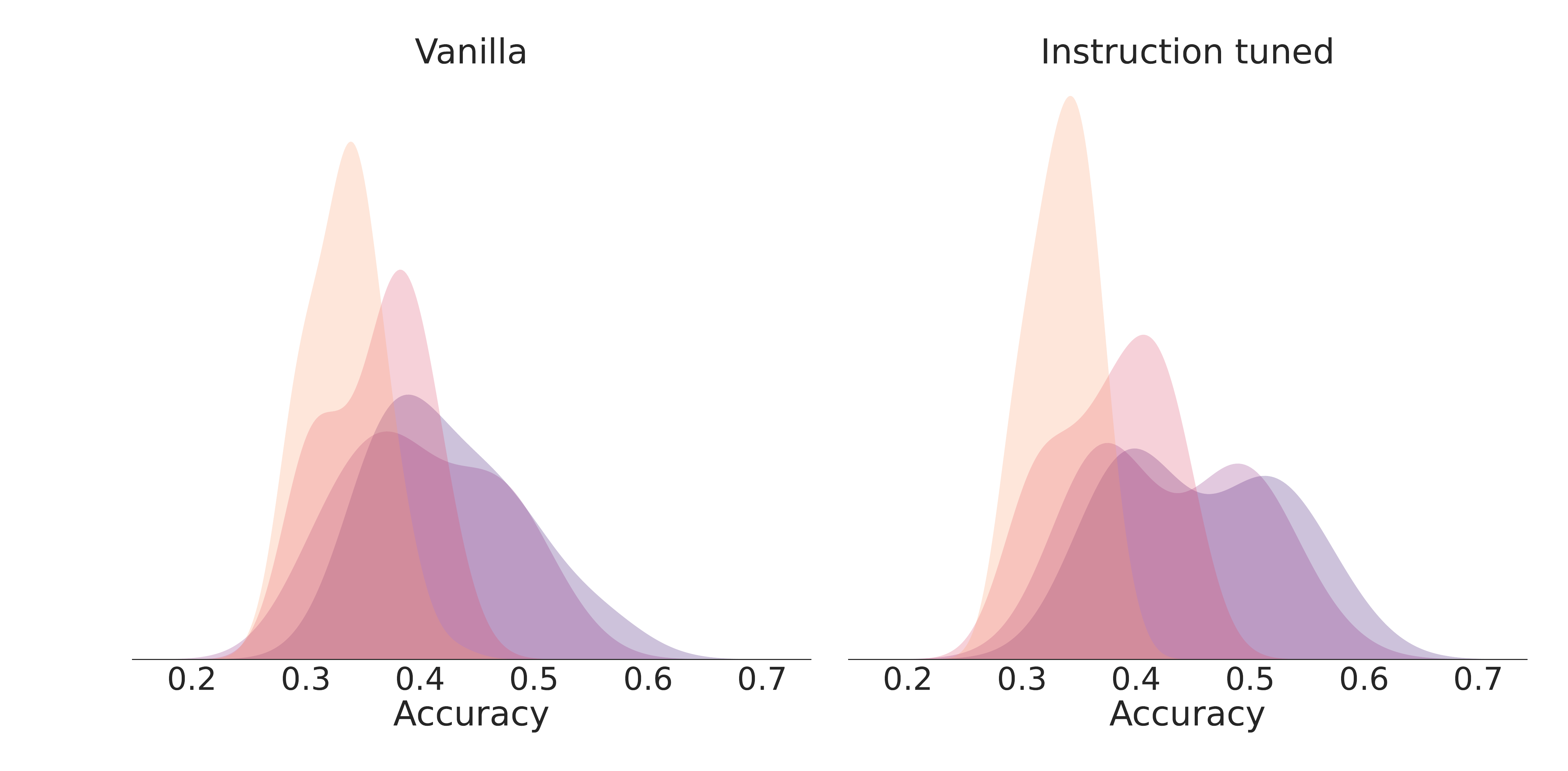}
    \caption{}
    \label{fig:distribution_results}
  \end{subfigure}

  \begin{subfigure}[h]{0.95\linewidth}
  \centering
    \includegraphics[width=\linewidth]{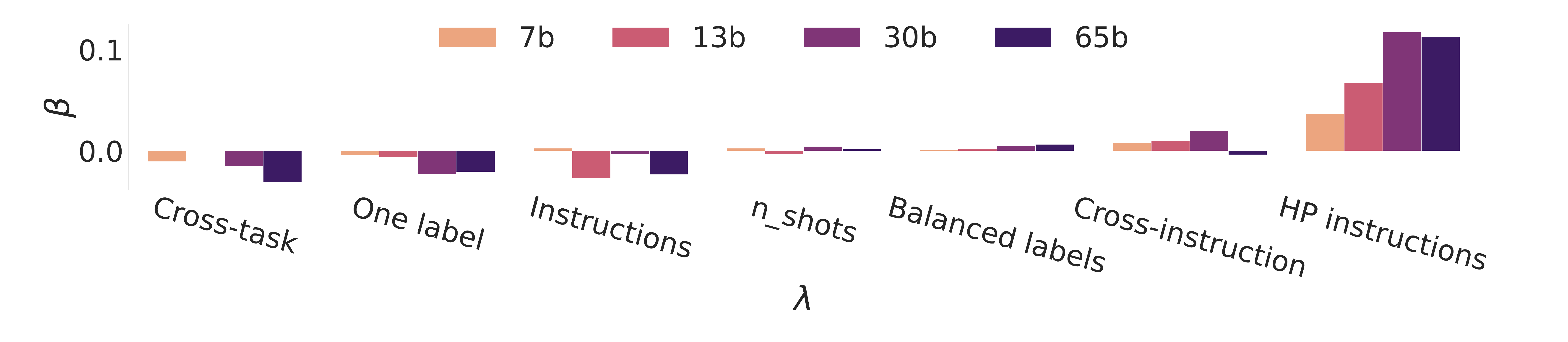}
    \caption{}
    \label{fig:main_effects}
  \end{subfigure}
    
  \caption{%
  The following performance estimates help to put the results of our consistency metric into perspective:
  (a) A kernel density estimation of the accuracy scores of all eight models across all of our setups.
  We see how the spread of accuracy scores is high, meaning that the accuracy of a model is very different depending on the setup in which it is evaluated.
  (b) The $\beta$-values of the main effects of each individual factor across many different runs. The values can be directly interpreted as `\emph{expected accuracy gain/loss}' when a factor is present compared to when it is absent.
  }
  \label{fig:results_acc_and_main_effects}
\end{figure*}

\paragraph{Main effects of $\lambda$}
Next to $\kappa$ -- the primary metric --, we also provide the auxiliary metric in the form of the main effects of factors. 
The main effects show how much the presence or absence of a factor influences the accuracy of the model on average. 
The main effects of the factors help to interpret their $\kappa$ values: Does the change in prediction occur because the factor actually improves the model accuracy, or is it due to model inconsistency?

To obtain measures of the main effects, we fit a simple linear regression model to predict accuracy scores from the presence or absence of each factor $Acc = \beta_1\lambda + \beta_0$.
We can then interpret the coefficient $\beta_1$ of $\lambda$ as its main effect ('How much does the factor on average change accuracy scores?').

\subsubsection{Selecting instructions} 
\label{subsubsec:probing_templates}
To find a set of high- and low-performing instructions for the \factor{HP instructions} factor, we run a preliminary analysis where we probe model behaviour in response to all 15 available P3 instructions for the ANLI dataset.
We assess the performance of different instructions based on accuracy and consistency.
A specification of the used models can be found in Table~\ref{tab:learning-models} and further details in Section~\ref{subsec:setup}.

\begin{table}[h!]
\centering
\footnotesize
\begin{tabular}{ll}
\hline
\textbf{Type of learning} & \textbf{Model}\\
\hline
\rule{0pt}{2ex}
ICL + vanilla 
 & LLaMA 7B, 13B, 30B, 65B \\
ICL + Instruction-tuning & Alpaca 7B, 13B, 30B, 65B\\
\hline
\end{tabular}
\caption{Models with their respective adaptation types, as used while selecting instructions (Section~\ref{subsubsec:probing_templates} and the empirical evaluation (Section~\ref{sec:experiments}.}
\label{tab:learning-models}
\end{table}

\begin{figure*}[h!]
  \centering
    \includegraphics[width=0.99\linewidth]{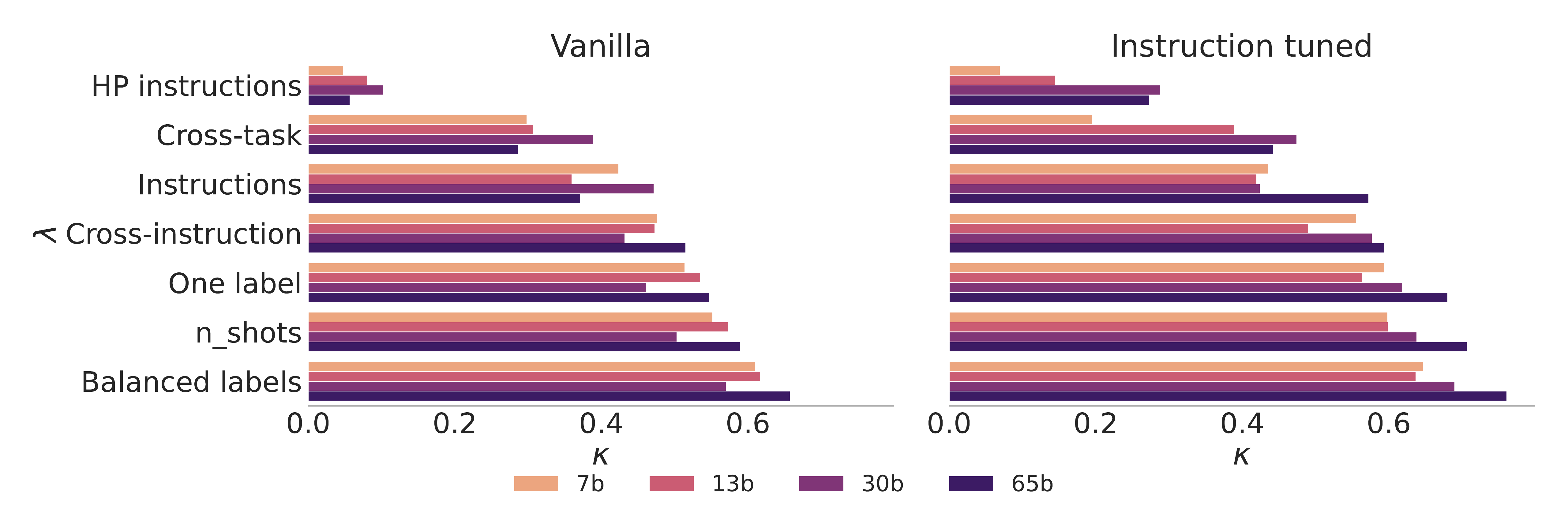}
  \caption{The consistency values comparing predictions when a specific factor $\lambda$ is present or not. A value of 0 indicates a complete change of predictions while a value of 1 indicates perfect agreement. Hence, a low value indicates that a model is not robust to a change in a specific $\lambda$.}
  \label{fig:consistency_lambdas}
\end{figure*}

We first get a general picture of each model's average consistency $\kappa_{avg}$ across all instructions.
We find that $\kappa_{avg}$ increases with the number of parameters and is overall higher when a model has been instruction tuned (Figure~\ref{subfig:consistency_templates_per_model}).

We then consider the consistency of each individual instruction and find a congruent pattern of consistency across all models (Figure~\ref{subfig:consistency_templates_per_template}) that corresponds generally to the accuracy scores of the same instructions (compare Figure~\ref{subfig:heatmap_performance_templates}).
Interestingly, we also find two groups of high-accuracy instructions making very different predictions (compare the consistency scores of 9, 10 and 15 vs. the rest).
Based on these observations, we choose the two highest- and lowest-performing instructions to constitute the \factor{HP instructions} factor and templates 14 and 15 as realisations of the \factor{instructions} factor.
Examples of the templates that we select to construct the ICL consistency test can be found in Appendix~\ref{app:template_examples}.

\section{Empirical evaluation}
\label{sec:experiments}
In the following section, we evaluate multiple LLMs of different scales on their robustness to setup changes.

\subsection{Models and sampling}
\label{subsec:setup}
We evaluate eight LLMs of different sizes and types of pretraining (for an overview, see Table~\ref{tab:learning-models}).
While `Vanilla' models are regular pre-trained LLMs \citep{touvron2023llama}, instruction-tuned models are the same models but additionally tuned to follow instructions \citep[see e.g.][]{wei2021finetuned, zhong2021adapting} via low-rank adaptation \citep[LoRA;][]{hu2021lora} on the alpaca self-instruct dataset \citep{alpaca, wang2022self}.
We run all models using mixed-precision decomposition as described by \citet{dettmers2022llm}. 

To obtain a prediction from our models, we greedily sample from their probability distribution over all possible labels, with the label space being determined by the respective instruction template.
However, constraining the sampling to the label distribution appears to be not strictly necessary. During our experiments, we observed that greedy sampling from the probability distribution over the whole vocabulary yields the same results.

\subsection{Results}
\label{subsec:results}
We evaluate all eight models on each setup using the full 600 data points of the ICL consistency test.
We add \factor{instruction tuning} as a custom factor to our analysis, following the description in Appendix~\ref{app:custom_factors}.
For our exemplary evaluation, we focus on the ANLI dataset.

\paragraph{Performance distribution}
Figure~\ref{fig:distribution_results} shows the distribution of accuracy scores across all runs for different models.
The spread of scores is strikingly wide, with the large models scoring from below chance to up to 67\% accuracy, depending on the setup.
This extreme variability underlines the importance of understanding the impact of different design decisions and prediction consistency in ICL. 

\paragraph{Main effects of $\lambda$}
Figure~\ref{fig:main_effects} presents the main effects of all factors separated by model size.
The $\beta$-values show us how a change in a factor influences the accuracy of a model across all setups in general.

We can see how the choice of instructions can largely improve performance (see \factor{HP instructions}).
Surprisingly, using varied instructions for the in-context examples (see \factor{Cross-instruction}) also slightly improves performance for 3 out of 4 models.
On the other end of the spectrum, using QQP in-context (\factor{cross-task}) as well as only providing in-context examples with only a single label (\factor{One label}) are generally deteriorating performance, and especially so for larger models.
\factor{Instructions} shows us that choosing the correct instruction template is difficult: The average gain and loss of accuracy is ambivalent, depending on the model that is tested.

These main effects give us a general idea of the tendencies of factors and help us understand the nature of potential prediction inconsistencies in $\kappa$ in the next section.

\paragraph{Model consistency}
Our primary metric, Cohen's $\kappa$, shows the consistency of a model's prediction for the same data point across many equivalent setups. 

First, we ensure that our consistency is not driven by a lack of diversity in model predictions \citep[imagine that our models always predict just a single label, as observed by e.g.][]{zhao2021calibrate}. 
This is not the case (for details, see Appendix~\ref{app:prediction_diversity}).
With this concern out of the way, we  first examine separated $\kappa$ for each factor $\lambda$. 
This measure tells us how much a model prediction changes across many setups if a specific factor is present compared to when it is absent. 
In Figure~\ref{fig:consistency_lambdas}, we see kappa scores for all $\lambda$s separated by model.  
Overall, model consistency is relatively poor: all models are susceptible to all factors, with the highest $\kappa$ value scoring at 0.75.

\begin{figure}[h!]
  \centering
    \includegraphics[width=0.99\linewidth]{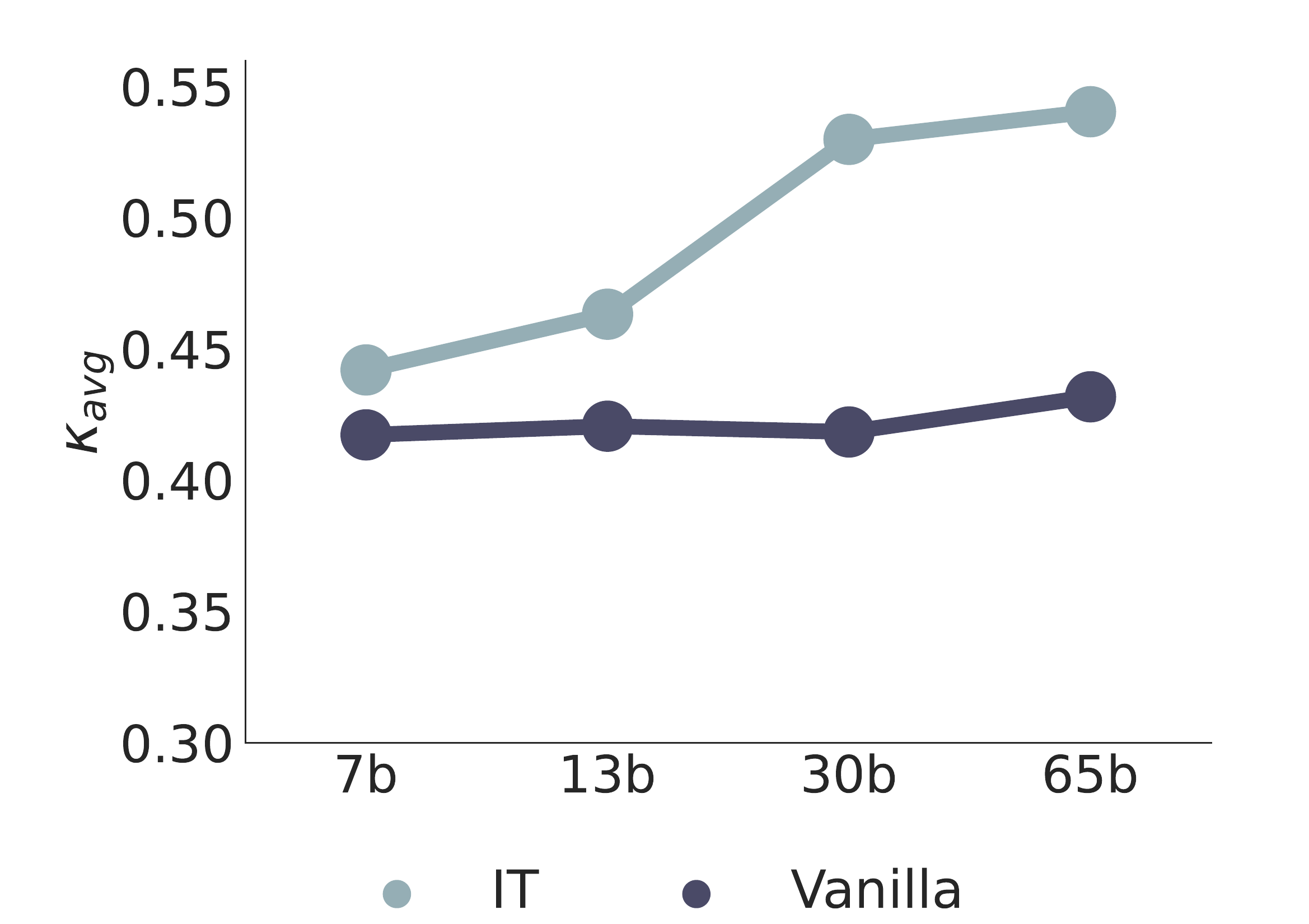}
  \caption{Average kappa per model. A high value reveals a model's robustness to setup changes.}
  \label{fig:consistency_avg}
\end{figure}

With their predictions being influenced by factors irrelevant to the task at hand, we can conclude that the predictions of all models appear to happen on non-robust generalisations. 
The predictions are not reliable.
Interestingly, robustness (or the lack thereof) to specific factors is highly correlated across models, with all models being more susceptible to the same factors (e.g. \factor{HP instructions}) while relatively robust to others (e.g. \factor{Balanced labels}). 

When we accumulate $\kappa$-scores, we again observe that IT models are generally more consistent in their predictions than vanilla models (see Figure~\ref{fig:consistency_avg}).
Interestingly, the consistency of a model improves for IT models with increasing parameter count, while additional parameters do not improve consistency for vanilla models. 
This indicates that instruction tuning is especially effective in improving the robustness of larger models.

\section{Conclusion}
\label{sec:conclusion}
In this paper, we have introduced a new test for robustness in prompt-based learning, the ICL consistency test.
The ICL consistency test evaluates the consistency of model predictions on the same data points across many different setups.

Our evaluation shows that the spread of accuracy scores across different setups is very high, indicating that model predictions are inconsistent and depend on the exact setup in which the models are evaluated. 
For example, the accuracy of the largest evaluated model on ANLI differs up to 40\% depending on the setup in which it is evaluated.
The primary metric $\kappa$ of the ICL consistency test showed that the models did not perform with high consistency for any minimal setup change (i.e. across any change in factors). 

The results suggest that the ICL consistency test is a good indicator of the quality of the generalisation an LLM is making: If predictions are consistent, the model correctly disregards irrelevant context information; if it is inconsistent, it lets irrelevant context information influence its predictions.
The fact that state-of-the-art LLMs score low on our test highlights important room for improvement in their generalisation capacities.

\section*{Limitations}
The presented ICL consistency test has several limitations in assessing model robustness. 
First and foremost, the test is currently only implemented for a single type of task (natural language inference), and model consistency might differ in other types of classification tasks or more open-ended answering formats such as question answering.
However, we think that the performance on the ICL consistency test can be a good indicator of the quality of the generalisation that an LLM is making: If predictions are not consistent, the model is influenced by irrelevant context information, and generalisation is therefore not robust.

Another limitation of the test is that the considered factors, albeit we think our choice of factors appropriate, are in no way exhaustive and additional factors might be informative. 
For this reason, we added the possibility to augment the test with user-defined factors. 
New factors can be added to the test seamlessly.

\bibliography{anthology,custom}
\bibliographystyle{acl_natbib}

\onecolumn
\appendix

\newpage

\section{Prompt template examples}
\label{app:template_examples}

In the following, we provide more details on the instruction templates \citep{bach2022promptsource} used to construct the ICL consistency test.

\subsection{P3 details -- names}
\label{subapp:p3_names}
Names of all available P3-instructions, ordered as in Figure~\ref{fig:diversity_predictions}
\begin{multicols}{3}
\begin{enumerate}
\small
\setlength\itemsep{0.1em}
\item `MNLI Crowdsource'
\item `Guaranteed Possible Impossible'
\item `Always Sometimes Never'
\item `Consider Always Sometimes Never'
\item `Does This Imply'
\item `Guaranteed True'
\item `GPT 3 Style'
\item `Take the Following as Truth'
\item `Must Be True'
\item `Based on the Previous Passage'
\item `Should Assume'
\item `Can We Infer'
\item `Justified in Saying'
\item `Does It Follow That'
\item `Claim True False Inconclusive'
\end{enumerate}
\end{multicols}

\subsection{P3 details -- examples}
\label{subapp:p3_examples}
\paragraph{High-performing templates}
`Claim true false inconclusive' \\
\fbox{\parbox{\linewidth}{
\textbf{[...]} \\

Jonathan Smith (born January 17, 1971), better known by his stage name Lil Jon, is an American rapper, record producer, and DJ. He was the frontman of the group Lil Jon \& The East Side Boyz, which he formed in 1997, and they released several albums until 2004. Based on that information, is the claim: "Jonathan Smith spent much of his time in China." true, false, or inconclusive? \\

ANSWER: 
}}
\paragraph{High-performing templates} `Does it follow that' \\
\fbox{\parbox{\linewidth}{
\textbf{[...]} \\
Given that Jonathan Smith (born January 17, 1971), better known by his stage name Lil Jon, is an American rapper, record producer, and DJ. He was the frontman of the group Lil Jon \& The East Side Boyz, which he formed in 1997, and they released several albums until 2004. Does it follow that Jonathan Smith spent much of his time in China. Yes, no, or maybe? \\

ANSWER: 
}}
\paragraph{Low-performing templates} `MNLI crowdsource' \\
\fbox{\parbox{\linewidth}{
\textbf{[...]} \\
Jonathan Smith (born January 17, 1971), better known by his stage name Lil Jon, is an American rapper, record producer, and DJ. He was the frontman of the group Lil Jon \& The East Side Boyz, which he formed in 1997, and they released several albums until 2004. Using only the above description and what you know about the world, "Jonathan Smith spent much of his time in China." is definitely correct, incorrect, or inconclusive? 

ANSWER: 
}}

\paragraph{Low-performing templates}
`Guaranteed possible impossible' \\
\fbox{\parbox{\linewidth}{
\textbf{[...]} \\
Assume it is true that Jonathan Smith (born January 17, 1971), better known by his stage name Lil Jon, is an American rapper, record producer, and DJ. He was the frontman of the group Lil Jon \& The East Side Boyz, which he formed in 1997, and they released several albums until 2004. \\

Therefore, "Jonathan Smith spent much of his time in China." is guaranteed, possible, or impossible? \\

ANSWER: 
}}

\section{Factor details}
\label{app:factors_details}
In the following, we provide a more detailed description of the factors shortly summarized in Section~\ref{subsec:setups_and_factors} and also offer our motivation to include these factors.

\paragraph{n-shots} The number of examples in-context is a factor that is always present in any in-context learning setting. 
As such, it is an essential factor to include, as it has high practical significance and should be considered with caution if it results in low $\kappa$ values.
We introduce few ($k$ = 2) and many ($k$ = 5) examples as a factor.

\paragraph{HP instructions} We have seen how some instructions produce consistent and relatively well-performing responses across different models while others do not (see Section~\ref{subsubsec:probing_templates}. 
We add this factor to increase the difficulty of the task: A robust model should be able to predict a consistent label in response to either of these equivalent instructions.
We chose the two best and two worst-performing templates\footnote{See Appendix~\ref{app:template_examples} for an example of the instructions} from our analysis in Section~\ref{subsubsec:probing_templates}.

\paragraph{Balanced labels} \citet{zhao2021calibrate} showed how a majority label in-context can influence the distribution of model outputs.
We, therefore, compare contexts with balanced in-context label distribution with randomly sampled labels.

\begin{figure*}[h!]
  \centering
    \includegraphics[width=\linewidth]{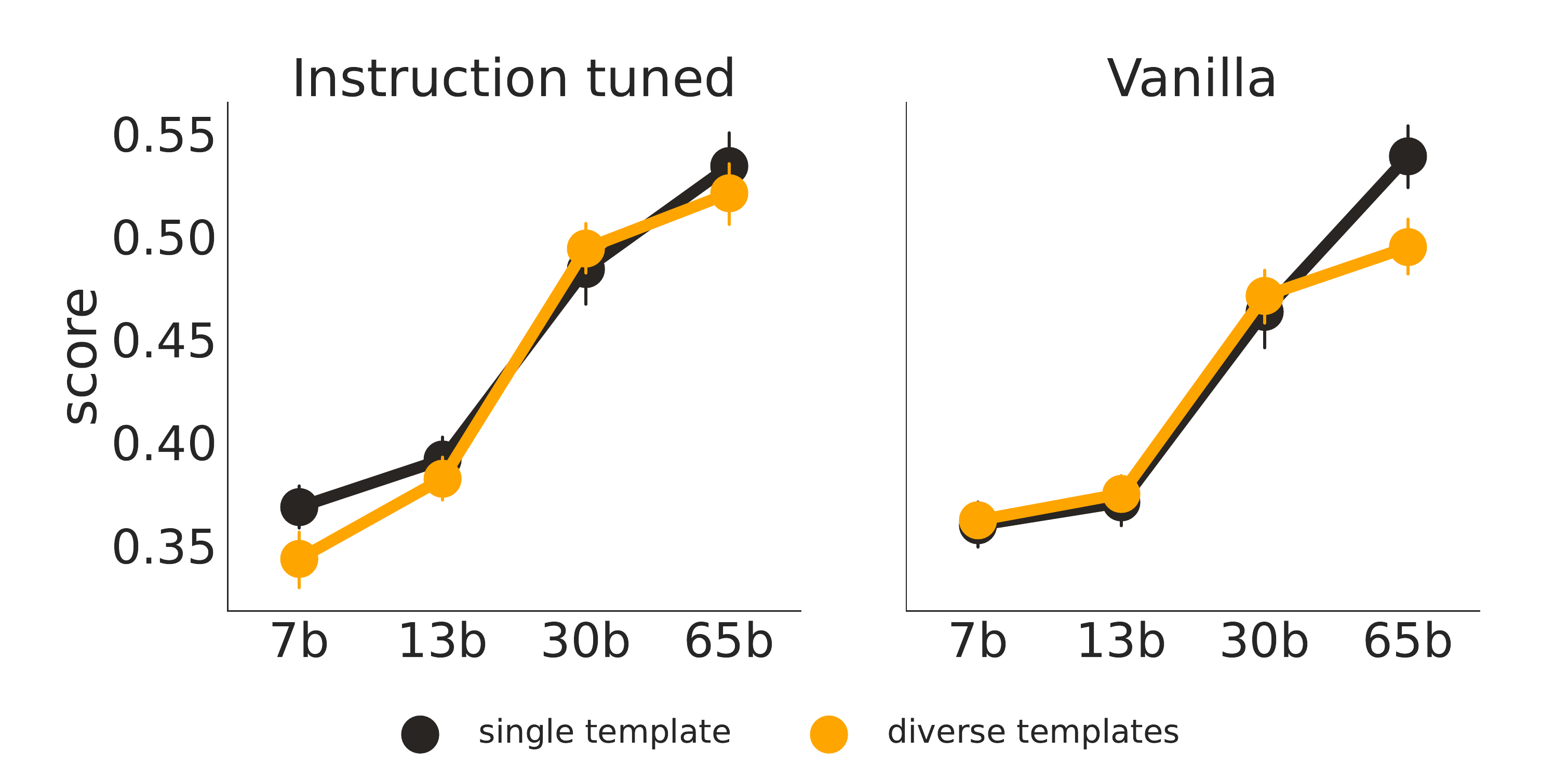}
  \caption{}
  \label{fig:diverse_templates_performance}
\end{figure*}

\paragraph{Cross-instruction}
We include cross-templates as a factor to assess model robustness to shifts in the label space $\mathcal{C}$ and surface form of instruction formulation.
Previous research has shown how in-context learners are sensitive to the instructions \citep{mishra2021reframing} as well as the label distribution $\mathcal{C}$ \citep{min2022rethinking}.
The experiments of \citet{min2022rethinking} represent an extreme case in which $\mathcal{C}$ is resampled as random tokens.
While these edge cases are theoretically attractive, we here change this scenario to a more common one, where instructions and labels are semantically equivalent but have different surface forms. 
We randomly sample from the set of available p3 instructions to obtain the in-context examples. 
Surprisingly, when testing this factor, we find that almost all models are robust to semantic-invariant changes to the in-context instructions (see Figure~\ref{fig:diverse_templates_performance}).
This happens despite changes in the label space and substantial changes in surface form and format across the used instructions. 

\begin{figure*}[h!]
  \centering
    \includegraphics[width=\linewidth]{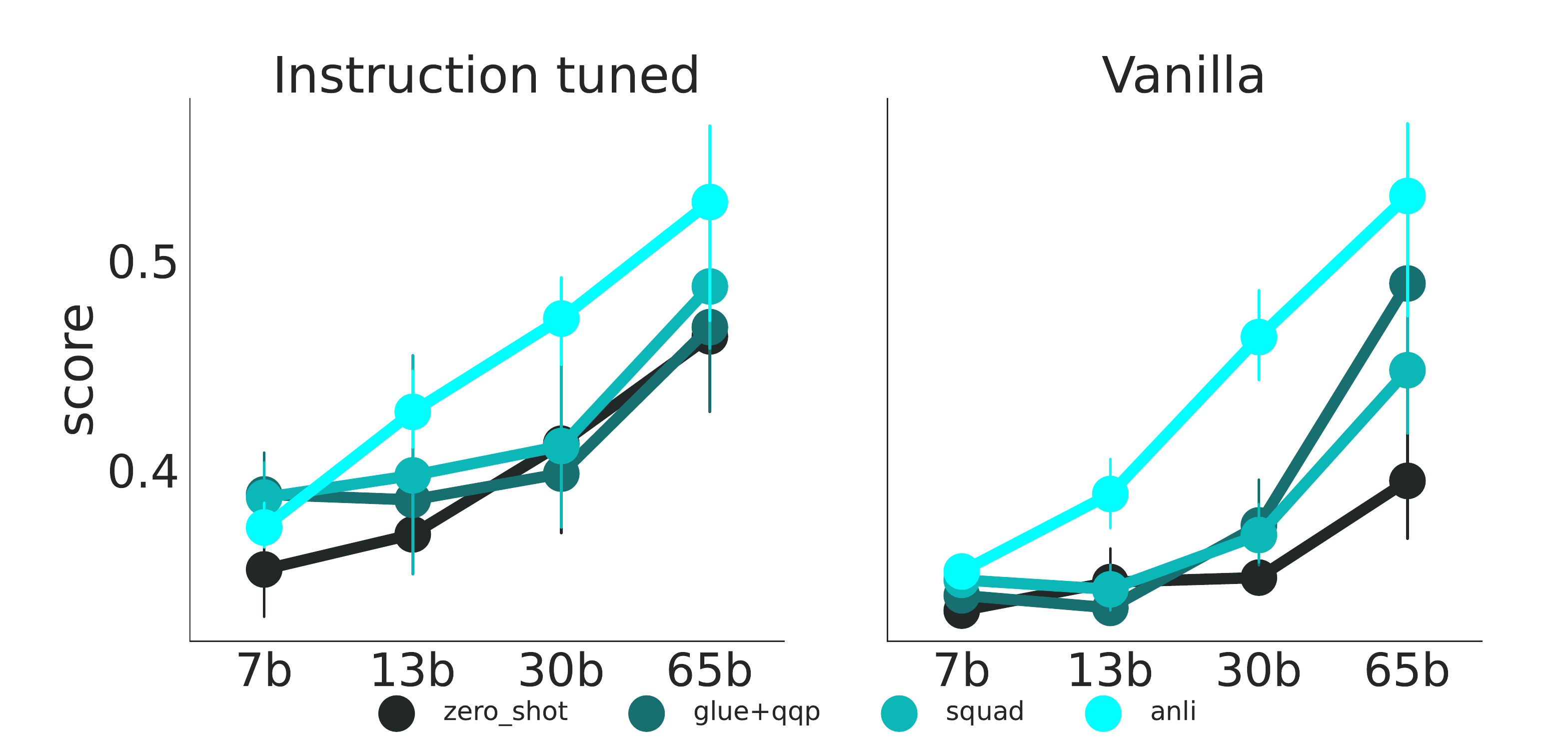}
  \caption{}
  \label{fig:cross_tasks}
\end{figure*}

\paragraph{Cross-task}
In \factor{cross-task}, we exchange the task of the in-context examples such that the only consistency between in-context and target examples is the general format ($x$ followed by $y$) and the truthfulness of the $x$ to $y$ mapping.
To see whether conditioning on a fixed label space matters, we add tasks with a discriminative (QQP) and a generative \citep[SQuAD;][]{rajpurkar2016squad} objective as different factors.
Compared to a zero-shot baseline, we can see that especially large models can benefit from conditioning on other tasks (Figure~\ref{fig:cross_tasks}).

For our test, we only include QQP as an in-context task, as SQuAD is incompatible with many other factors (such as \factor{balanced labels}, \factor{one label} aso...)

\paragraph{Instructions} Besides the performance of the instructions, we are also interested in how consistent model behaviour is across instructions with similar performance. 
We bin the two high-performing instructions into a new factor to get insights into this.


\section{Introducing custom factors}
\label{app:custom_factors}
The ICL consistency test allows the addition of additional user-defined factors.
This is useful if factors should be evaluated that are related to modifications of the model \citep[e.g. whether it was instruction-tuned][or not]{wei2021finetuned} or when the model was evaluated in a different way \citep[e.g. whether we calibrate our output probabilities][or not]{zhao2021calibrate}.
Note that adding a factor in this way will change the overall results of the analysis (see Section~\ref{subsec:metrics} for more details). 
Alternatively, the task can be evaluated separately for either user-defined factors.

\section{Details setup\_IDs}
\label{app:setup_IDs}
\begin{figure}[h!]
  \centering
    \includegraphics[width=\linewidth]{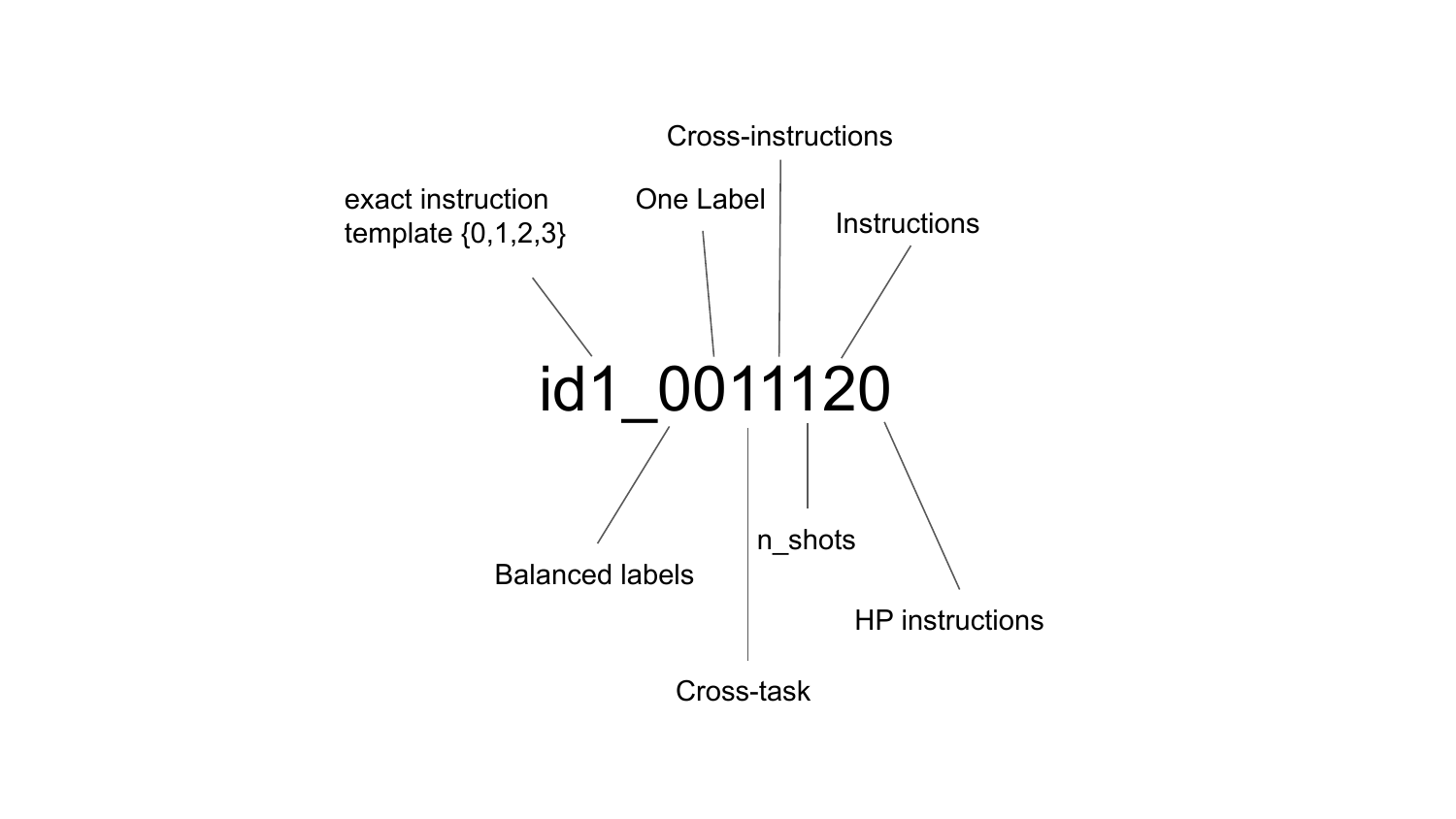}
  \caption{An illustration of the meaning of every digit in a setup\_ID: A factor is either absent (0), present (1) or irrelevant for the setup (2).}
  \label{fig:setup_ID_explanation}
\end{figure}

Setups are defined by the presence and absence of factors. 
The setup\_IDs reflect how the factors make up the setup (see Figure~\ref{fig:setup_ID_explanation}). 
Setup\_IDs can also be converted into a human-readable format by using the `\texttt{\_convert\_numeric\_id\_to\_dict}'-method of the `\texttt{task}'-object in the code submission to the GenBench CBT.

\section{Prediction diversity}
\label{app:prediction_diversity}
\begin{figure}[h!]
  \centering
    \includegraphics[width=\linewidth]{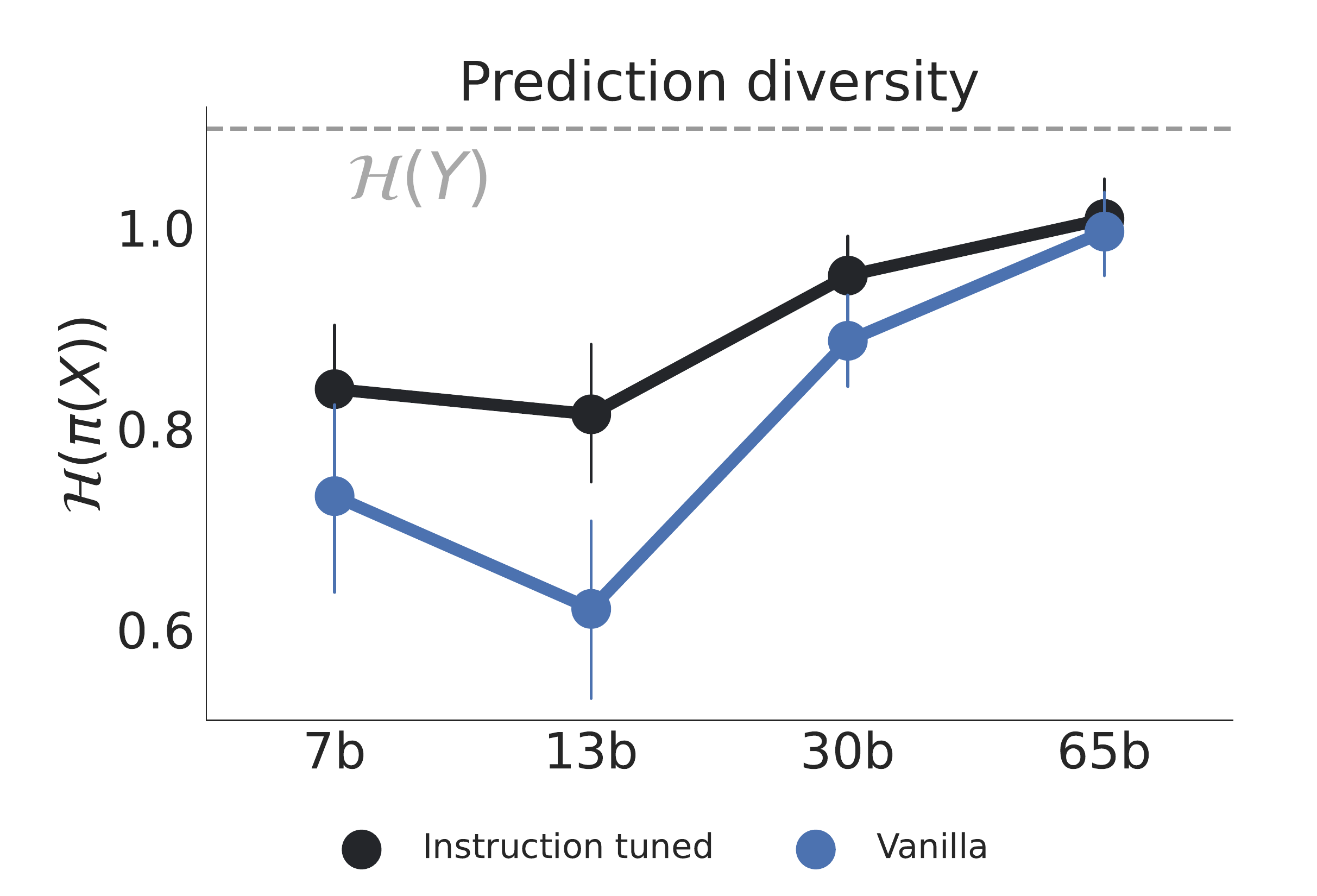}
  \caption{The diversity of model predictions across the whole dataset measured through entropy. A value close to 0 indicates that a model is strongly biased to always predict the same label. A value close to the entropy of the true label distribution $\mathcal{H}(Y)$ indicates that the model has no bias towards any specific label.}
  \label{fig:entropy_label_bias}
\end{figure}

Models could achieve a high consistency score by always predicting the same label. 
We check whether models tend to do so by calculating the entropy of a model's predictions across all data points in the ICL consistency test. 
This allows us to estimate whether a model is biased toward predicting a single label (low entropy). 
An unbiased model's prediction should be close to the entropy of the target distribution $\mathcal{H}(Y)$. 
We find that smaller models have a larger bias towards predicting a single label (lower prediction entropy), while larger and IT models get closer to $\mathcal{H}(Y)$ (see Figure~\ref{fig:entropy_label_bias}).
This hints that we might overestimate the $\kappa$ values for smaller vanilla models.

\end{document}